\documentclass[11pt]{article}

\usepackage[preprint]{acl}

\usepackage{times}
\usepackage{latexsym}
\usepackage{booktabs}
\usepackage{amsmath}
\usepackage{amssymb}
\usepackage{multirow}
\usepackage{xcolor}
\usepackage{url}
\usepackage{array}
\usepackage{graphicx}
\usepackage{placeins}
\usepackage{stfloats}

\usepackage[T1]{fontenc}

\usepackage[utf8]{inputenc}

\usepackage{microtype}

\usepackage{inconsolata}

\usepackage{graphicx}

\title{Hi-Token: Hierarchical Coordinate Tokenization for \\ Generative Visual Grounding}

\author{
 \textbf{Xiuyuan Zhu\textsuperscript{1,2}},
 \textbf{Ke Lu\textsuperscript{1,3}},
 \textbf{Kun Dong\textsuperscript{1,2}},
 \textbf{Siwen Jiao\textsuperscript{4}},
\\
 \textbf{Hao Wu\textsuperscript{1,2}},
 \textbf{Zijin Du\textsuperscript{1}},
 \textbf{Shun Mao\textsuperscript{1,2}},
 \textbf{Dongming Zhang\textsuperscript{2}},
 \textbf{Jian Xue\textsuperscript{1,*}}
\\
\\
 \textsuperscript{1}University of Chinese Academy of Sciences, Beijing, China
\\
 \textsuperscript{2}State Key Laboratory of Communication Content Cognition, Beijing, China
\\
 \textsuperscript{3}Peng Cheng Laboratory, Shenzhen, Guangdong, China
\\
 \textsuperscript{4}National University of Singapore, Singapore
\\
 \small{
   \textsuperscript{*}Correspondence: xuejian@ucas.ac.cn
 }
}

\begin{document}
\maketitle
\begin{abstract}
Generative Vision-Language Models (VLMs) express bounding boxes as coordinate sequences, whose token identities often leave axis and place value implicit. We study how explicitly encoding these roles improves visual grounding. Hi-Token assigns a distinct token to each combination of axis, decimal place, and digit value, representing a box with 12 tokens drawn from 60 types. This design exposes coarse-to-fine coordinate structure and increases token reuse relative to atomic location tokens while retaining the existing VLM architecture. Hi-GAR complements this representation with a geometry-based reward for Group Relative Policy Optimization (GRPO), combining box overlap, coordinate accuracy at multiple scales, strict-IoU bonuses, and validity gating. Matched training comparisons show that Hi-Token improves localization throughout the evaluated IoU range, with gains across three VLM backbones. Hi-Token also outperforms an additionally tuned atomic-token baseline on the RefCOCO family before reinforcement learning. Hi-GAR further improves localization and is used only during training. The resulting Hi-R1 model achieves higher values than strong specialist baselines on most reported metrics. Token-frequency and geometric diagnostics characterize the representation and reward stages. These results establish explicit axis and place-value encoding as an effective design for generative visual grounding. 
\href{https://xyzzzh.github.io/Hi-Token/}{\textcolor{blue}{Project page}}.
\end{abstract}

\section{Introduction}
\label{sec:intro}

\begin{figure}[t]
  \centering
  \includegraphics[width=\linewidth, trim=10 100 435 10, clip]{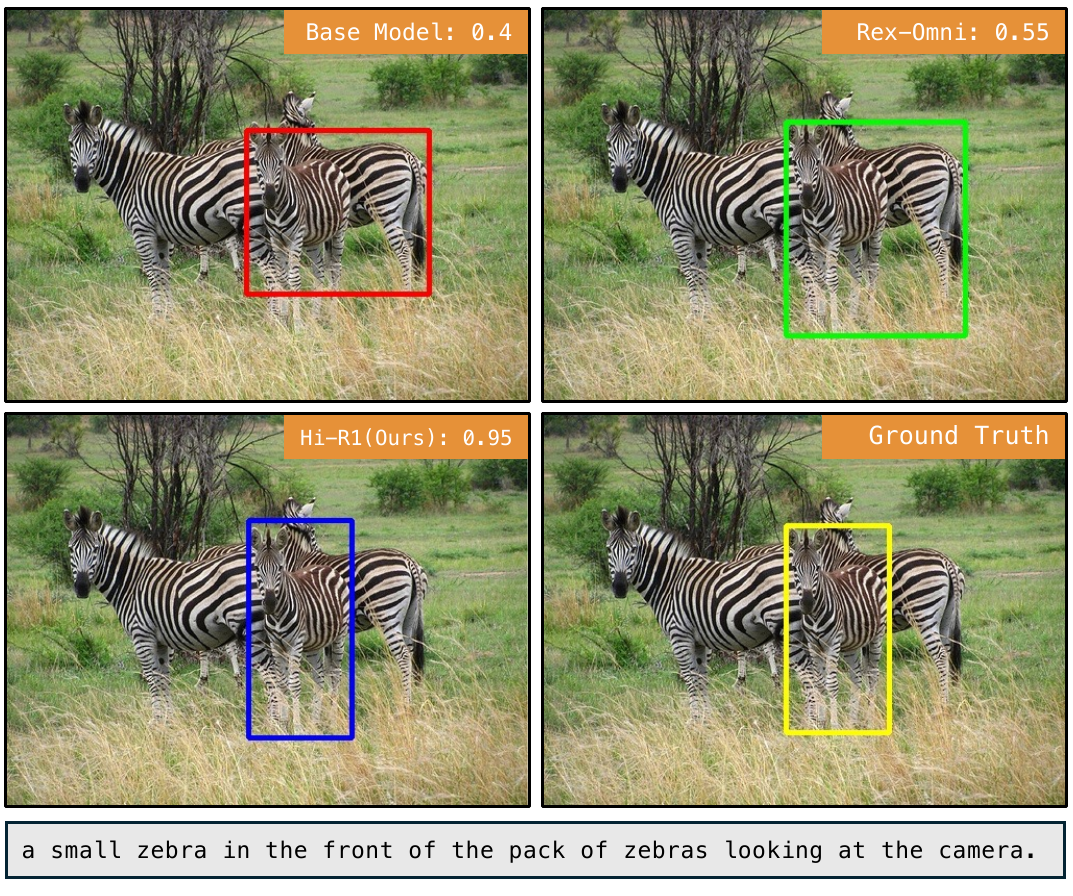}
  \caption{\textbf{Qualitative comparison.}
  For a query with overlapping targets, Qwen2.5-VL-7B and Rex-Omni produce less accurate boxes with IoU scores of 0.40 and 0.55, respectively.
  The full Hi-R1 model localizes the referred zebra more accurately, reaching an IoU of 0.95.}
  \label{fig:qualitative_zebra}
\end{figure}

Visual grounding connects linguistic expressions to image regions and is a basic capability of multimodal assistants and embodied systems. Vision-Language Models (VLMs)~\cite{gpt5,gemini,qwenvl,qwen2.5vl,llava1,llava2,llava3} provide a text-generation interface in which bounding boxes can be represented as sequences of coordinate tokens. Performance across IoU thresholds reflects both target identification and boundary alignment. Figure~\ref{fig:qualitative_zebra} illustrates the distinction. For a query referring to one zebra in a group, existing generative models locate the relevant area but produce boxes that include adjacent animals or miss part of the target. Reliable grounding therefore depends on the model's ability to generate a geometrically coherent coordinate sequence after it has understood the query and image.

Coordinate representation directly shapes this generation problem. Existing methods express boxes as numeric coordinate strings~\cite{shikra,vlm-r1,qwenvl,qwen2.5vl,qwen3vl,seed1.5} or atomic location tokens~\cite{pix2seq,rex-omni}. Numeric strings can already be generated digit by digit: the base tokenizer used here splits a three-digit coordinate into three digit tokens. These digit identities are shared across axes and decimal places, so the model infers each digit's role from context. Atomic location tokens assign a separate identity to each complete coordinate, distributing training observations over 1,000 types. Nearby values such as \texttt{\textless 323\textgreater} and \texttt{\textless 324\textgreater} therefore share no digit tokens. These designs motivate a coordinate vocabulary that combines reusable digits with explicit axis and place-value roles.

Our goal is to make coordinate generation easier to learn while preserving the standard autoregressive VLM architecture and an exactly parseable output format. We focus on two aspects of coordinate generation. First, the representation must expose numerical scale and axis roles while remaining compact. Second, the training signal must reflect the geometric effect of coordinate errors. Supervised fine-tuning assigns likelihood to individual tokens, although equal token errors can produce different changes in box overlap. A box-level IoU reward provides useful global feedback but gives limited information about coordinate accuracy at different scales. 

We address the representation problem with Hi-Token. Each token identifies an axis, a decimal place, and a digit value. A box is represented by 12 tokens drawn from $2\times3\times10=60$ coordinate types. This vocabulary gives digits explicit spatial roles while sharing them across coordinate values with the same axis and place. Relative to 1,000 atomic location tokens, it concentrates supervision over fewer types. Hi-Token changes only the output vocabulary and sequence. We address the training problem with Hi-GAR, a reward for Group Relative Policy Optimization (GRPO)~\cite{grpo1,grpo2}. Hi-GAR combines box IoU with coordinate checks at several tolerances and bonuses at multiple IoU thresholds. A validity gate disables coordinate-level rewards when the predicted box has negligible overlap with the target. Hi-GAR is used only during training. We refer to the model trained with both components as Hi-R1.

We evaluate the representation and reward stages separately. Under the same backbone, data, optimization, decoding, and evaluator, Hi-Token improves RefCOCO mIoU, P@0.5, and P@0.95 by 14.1, 14.9, and 8.7 points over atomic coordinate tokens. After additional Flat-specific tuning, Hi-Token retains P@0.95 advantages of 5.4, 4.2, and 15.9 points on RefCOCO, RefCOCO+, and RefCOCOg, respectively. Cross-backbone comparisons on two Qwen3-VL models also favor Hi-Token. Token-frequency analysis characterizes the reuse and coverage of the complete encoding. With Hi-Token fixed, Hi-GAR improves localization over IoU-only GRPO, and a gate ablation measures the benefit of suppressing rewards for near-zero-overlap boxes. Hi-R1 obtains the highest value in eight of the nine benchmark columns reported for the RefCOCO family, and its inference latency remains close to that of the SFT model.

Our contributions are summarized as follows:
\begin{enumerate}
    \item We introduce Hi-Token, a coordinate representation whose token identities explicitly encode axis, decimal place, and digit value, supporting coarse-to-fine generation within a standard autoregressive VLM.

    \item We introduce Hi-GAR, a training reward that combines box overlap and coordinate accuracy at multiple scales. Its validity gate suppresses coordinate-level rewards for near-zero-overlap boxes.

    \item Matched comparisons show that Hi-Token improves localization across the evaluated IoU thresholds, with gains in P@0.95 across three VLM backbones. Token-frequency and diagnostic results characterize supervision density, digit boundaries, object-scale effects, and the separate contribution of reward training.
\end{enumerate}

\begin{figure*}[!ht]
    \centering
    \includegraphics[width=\textwidth,trim=5 50 10 90,clip]{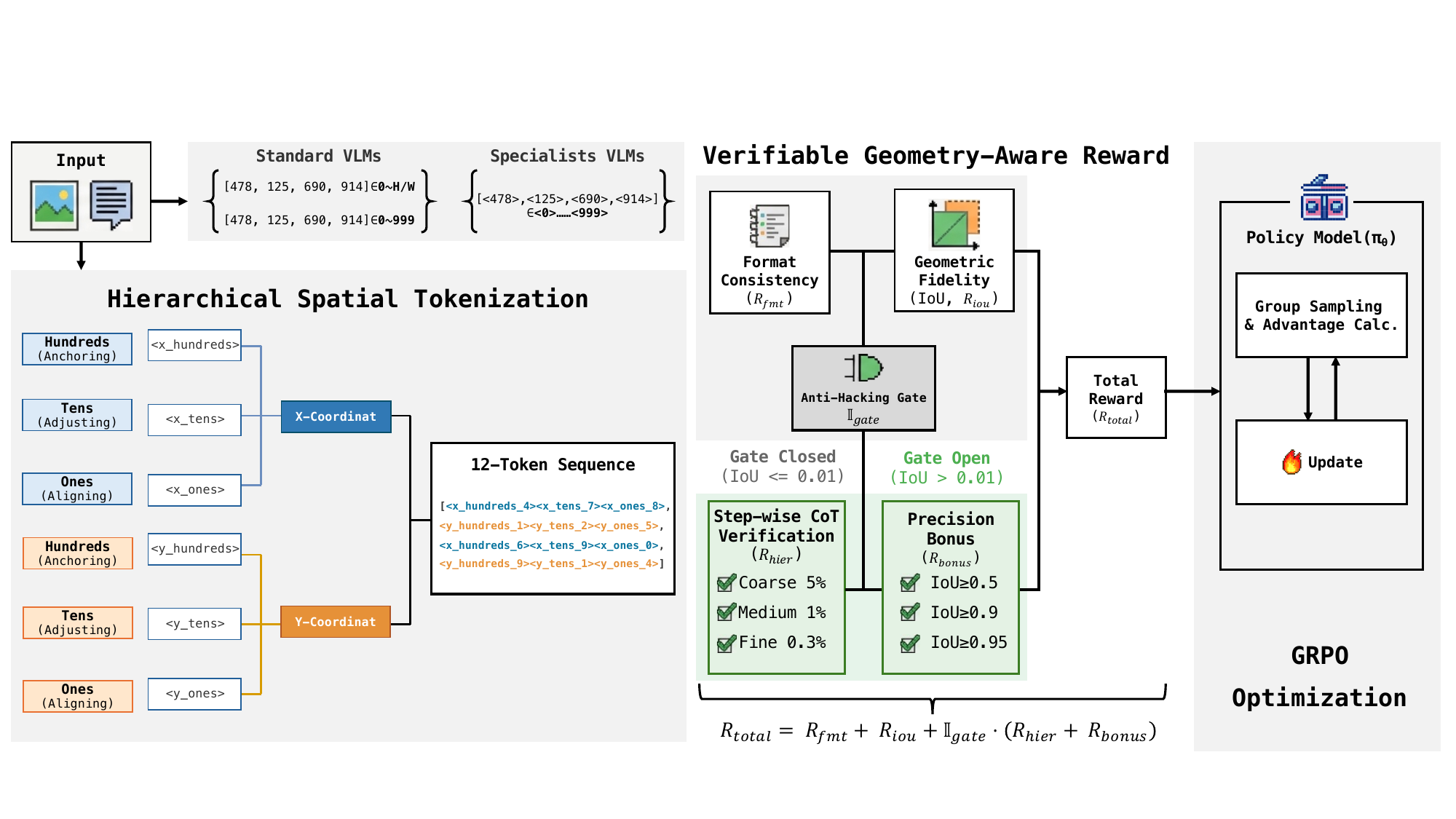}
    \caption{\textbf{Overview of Hi-Token and the geometry-aware post-training framework.}
    \textbf{Left:} Hi-Token represents each bounding box as a 12-token sequence by decomposing each coordinate into axis-specific hundreds, tens, and ones tokens.
    \textbf{Right:} The model is optimized with GRPO using the Hi-GAR reward.
    The reward combines format validity, IoU feedback, tiered coordinate verification, and strict-IoU bonuses:
    $R_{\mathrm{total}} = \lambda_{\mathrm{fmt}}R_{\mathrm{fmt}} + \lambda_{\mathrm{iou}}R_{\mathrm{iou}} + \mathbb{I}_{\mathrm{gate}}  \left( \lambda_{\mathrm{hier}}R_{\mathrm{hier}} + \lambda_{\mathrm{bonus}}R_{\mathrm{bonus}} \right)$.
    The validity gate $\mathbb{I}_{\mathrm{gate}}$ applies coordinate-level rewards only when the predicted box has sufficient overlap with the target.}
  \label{fig:framework}
\end{figure*}

\section{Related Work}
\label{sec:related}

\paragraph{Visual grounding paradigms.}
Visual grounding methods are commonly divided into regression-based and generation-based approaches.
Regression-based methods, such as DETR~\cite{detr}, Grounding DINO~\cite{grounding-dino}, OWLv2~\cite{owlv2}, and YOLO-World~\cite{yoloworld}, predict boxes with task-specific localization heads.
They often provide accurate localization, but are less flexible than general-purpose VLMs for open-ended multimodal interaction.
Generation-based methods, including Shikra~\cite{shikra}, Kosmos~\cite{kosmos}, Qwen-VL~\cite{qwenvl}, Qwen2.5-VL~\cite{qwen2.5vl}, and LLaVA-style models~\cite{llava1,llava2,llava3}, formulate localization as autoregressive sequence generation.
This formulation unifies grounding with instruction following, but recent studies~\cite{rex-omni} show that such models remain limited under strict localization metrics.
Hi-Token follows the generation-based paradigm and improves its coordinate representation.

\paragraph{Coordinate representation and supervision.}
Generative grounding methods use numeric coordinates~\cite{shikra,vlm-r1,qwenvl,qwen2.5vl,qwen3vl,seed1.5} or dedicated location tokens~\cite{pix2seq,rex-omni}.
Our base tokenizer splits a three-digit coordinate into three digit tokens, whose identities are shared across axes and decimal places.
The model therefore infers each digit's axis and place value from context.
Atomic location tokens assign a separate identity to each quantized coordinate value.
At the supervision level, DIST$^2$Loss~\cite{chung2026teaching} constructs distance-aware token targets and adds a KL-divergence term to cross-entropy training.
Hi-Token explicitly encodes axis, decimal place, and digit value in each token, yielding $2\times3\times10=60$ coordinate-token types.
The hundreds, tens, and ones tokens specify successive levels of precision, while separate axis identities distinguish horizontal and vertical coordinates.
This design combines explicit spatial roles with increased token reuse relative to the 1,000-type atomic representation~\cite{rex-omni}.
Matched SFT comparisons with the atomic-token baseline show gains in strict localization across three VLM backbones.

\paragraph{Reinforcement learning for grounding.}
Following verifiable rewards and GRPO~\cite{grpo1,grpo2}, VLM-R1~\cite{vlm-r1}, Visual-RFT~\cite{virft}, Perception-R1~\cite{perception-r1}, VisionReasoner~\cite{visionreasoner}, UniVG-R1~\cite{univg-r1}, No-Thinking-RL~\cite{no-thinking-rl}, DeepGrounder~\cite{zhang2026deepgrounder}, and Smooth Operator~\cite{jiao2026smooth} apply reinforcement learning to perception and localization.
In single-box grounding, VLM-R1 combines IoU and format rewards, while Perception-R1 uses IoU-based feedback, with squared IoU in its released grounding implementation.
Both retain native numeric coordinate outputs.
Hi-Token supplies explicit axis and place identities, and Hi-GAR adds coordinate-error checks at 50, 10, and 3 bins, IoU milestone bonuses, and a gate that suppresses auxiliary coordinate rewards for nearly non-overlapping boxes.
These terms provide feedback on individual coordinates alongside overall box overlap.
With Hi-Token fixed, our IoU-only and gate ablations assess the resulting reward refinement without adding inference-time modules.

\section{Method}
\label{sec:method}

\subsection{Overview}
\label{subsec:method_overview}

Hi-Token aims to improve the coordinate representation used by generative visual grounding models.
Existing autoregressive VLMs usually represent a coordinate as either a numeric string or an atomic location token.
Hi-Token makes axis and decimal place explicit in each digit token, combining numerical decomposition with dedicated spatial roles.
This section introduces the representation and Hi-GAR, a geometry-aware GRPO reward used to refine its predictions.

As shown in Figure~\ref{fig:framework}, our method has two components.
First, \textbf{Hi-Token} represents each coordinate with three axis-specific tokens corresponding to hundreds, tens, and ones digits.
This converts each bounding box into a 12-token sequence and introduces a coarse-to-fine structure into coordinate generation.
Second, \textbf{Hi-GAR} provides geometry-aware post-training feedback through box-level IoU, tiered coordinate verification, strict-IoU bonuses, and validity gating.
We denote the final model trained with both Hi-Token and Hi-GAR as \textbf{Hi-R1}.

\subsection{Hierarchical Coordinate Tokenization}
\label{subsec:hi_token}

\paragraph{Motivation.}
Flat coordinate tokenization treats each coordinate as an atomic category.
For example, in a flat location-token vocabulary, \texttt{<323>} and \texttt{<350>} are two independent tokens, although the corresponding coordinates are spatially close.
In addition, a shared coordinate vocabulary for both $x$ and $y$ axes does not explicitly encode axis-specific spatial patterns.
Hi-Token addresses these issues by decomposing each coordinate into a small sequence of axis-specific digit tokens.

\paragraph{Axis-decoupled coordinate vocabularies.}
We define a bounding box as
$B=[x_{\min}, y_{\min}, x_{\max}, y_{\max}]$ in the normalized coordinate space $[0,1]^2$.
For each axis $a \in \{x,y\}$, we construct a separate token vocabulary $\mathcal{V}_a$.
Each vocabulary contains digit tokens for three levels: hundreds, tens, and ones.
This gives the model separate token spaces for horizontal and vertical coordinates.

Given a continuous coordinate $v \in [0,1]$ on axis $a$, we first quantize it into an integer index:
\begin{equation}
    I_v = \left\lfloor v(M-1) \right\rfloor,
    \qquad M=1000 .
\end{equation}
We then decompose $I_v$ into three digits:
\begin{equation}
    \tau_v^a = (w_h^a, w_t^a, w_o^a),
\end{equation}
where $w_h^a$, $w_t^a$, and $w_o^a$ denote the hundreds, tens, and ones tokens on axis $a$, respectively.
The corresponding normalized coordinate is reconstructed as:
\begin{equation}
\begin{aligned}
    \hat{v}
    = \frac{
    100 \cdot \mathrm{val}(w_h^a)
    + 10 \cdot \mathrm{val}(w_t^a)
    + \mathrm{val}(w_o^a)
    }{M-1}.
\end{aligned}
\end{equation}
where $\mathrm{val}(\cdot)$ extracts the digit value from a token.
For example, $\mathrm{val}(\texttt{<x\_tens\_7>})=7$.

\paragraph{Bounding-box sequence.}
A bounding box is represented by concatenating the hierarchical tokens of its four coordinates:
\begin{equation}
\begin{aligned}
    \mathcal{S}_{\mathrm{box}} = [
    &\tau_{x_{\min}}^x,\,
     \tau_{y_{\min}}^y,\,
     \tau_{x_{\max}}^x,\,
     \tau_{y_{\max}}^y
    ] .
\end{aligned}
\end{equation}
Since each $\tau$ contains three tokens, the full box sequence contains 12 tokens.
For example, the point $(102,71)$ is tokenized as:
\begin{center}
\begin{tabular}{l}
\texttt{[<x\_hundreds\_1><x\_tens\_0><x\_ones\_2>,}\\
\texttt{\ <y\_hundreds\_0><y\_tens\_7><y\_ones\_1>]} .
\end{tabular}
\end{center}

\paragraph{Coarse-to-fine coordinate generation.}
Hi-Token factorizes coordinate generation into three ordered decisions:
\begin{equation}
\begin{aligned}
    P(\tau_v^a \mid I,T)
    &= P(w_h^a \mid I,T) \\
    &\quad \cdot P(w_t^a \mid w_h^a,I,T) \\
    &\quad \cdot P(w_o^a \mid w_h^a,w_t^a,I,T),
\end{aligned}
\end{equation}
where $I$ and $T$ denote the input image and text query.
For a quantized $x$-coordinate of $478$, the hundreds token selects
the interval $[400,499]$, the tens token narrows it to $[470,479]$,
and the ones token specifies $478$.
Coordinates from $470$ to $479$ share the first two tokens,
so training examples with different coordinate values provide
supervision for common axis- and place-specific tokens.
Each generated prefix thus has a concrete spatial meaning,
while shared tokens pool supervision across coordinate values.
The same value on the $y$ axis uses a separate token sequence.
This coarse-to-fine representation is generated by the standard
autoregressive decoder.

\subsection{Geometry-Aware GRPO Training}
\label{subsec:training}

Hi-Token improves the representation of bounding boxes, but standard supervised fine-tuning (SFT) still optimizes token likelihood rather than geometric quality.
A prediction with a small coordinate error and a prediction with a large coordinate error can receive similar token-level penalties if both differ from the target token.
We therefore use reinforcement learning to provide direct geometric feedback after SFT.

We adopt Group Relative Policy Optimization (GRPO)~\cite{grpo1,grpo2}.
For a prompt $q=(I,T)$, the policy samples $G$ candidate box sequences.
Hi-GAR assigns each sequence a scalar reward, which is standardized within the group to obtain its advantage.
The policy is updated with a clipped probability-ratio objective and KL regularization toward the reference policy.
Appendix~\ref{appendix:implementation} provides the training configuration.

\subsection{Geometry-Aware Reward Design}
\label{subsec:reward}

Hi-GAR evaluates the decoded coordinates at several geometric scales.
It combines box-level overlap, coordinate-level accuracy, strict-IoU milestones, and a validity gate. The geometric terms are computed from reconstructed boxes; the format term checks Hi-Token output syntax.
The total reward is defined as:
\begin{equation}
\begin{aligned}
R_{\mathrm{total}}
= &\lambda_{\mathrm{fmt}} R_{\mathrm{fmt}}
+ \lambda_{\mathrm{iou}} R_{\mathrm{iou}}  \\
&+ \mathbb{I}_{\mathrm{gate}}
\left(
\lambda_{\mathrm{hier}} R_{\mathrm{hier}}
+ \lambda_{\mathrm{bonus}} R_{\mathrm{bonus}}
\right).
\end{aligned}
\end{equation}
The gate $\mathbb{I}_{\mathrm{gate}}$ activates coordinate-level rewards only when the predicted box has sufficient overlap with the target.

\paragraph{Format reward.}
$R_{\mathrm{fmt}}$ checks whether the parsed answer contains exactly 12 coordinate tokens of the form \texttt{<axis\_level\_digit>}, with an $x$ or $y$ axis prefix and a hundreds, tens, or ones level.
It is defined as:
\begin{equation}
R_{\mathrm{fmt}} =
\begin{cases}
1, & \text{if the output format is valid}, \\
0, & \text{otherwise}.
\end{cases}
\end{equation}
The parser reconstructs four coordinates from the digit values of consecutive token triples.
Unparseable outputs receive zero total reward.

\paragraph{IoU reward.}
$R_{\mathrm{iou}}$ is the standard Intersection over Union between the predicted box and the ground-truth box:
\begin{equation}
    R_{\mathrm{iou}} = \mathrm{IoU}(B_{\mathrm{pred}}, B_{\mathrm{gt}}).
\end{equation}
It provides a continuous box-level reward for localization quality.
Predictions with reversed corner coordinates or negligible union area receive zero IoU reward.

\paragraph{Validity gate.}
Coordinate-level rewards can be misleading when the predicted box has almost no overlap with the target.
For example, a degenerate prediction may match one corner coordinate but fail to form a valid target box.
We therefore use a validity gate:
\begin{equation}
    \mathbb{I}_{\mathrm{gate}}
    =
    \mathbb{I}
    \left(
    \mathrm{IoU}(B_{\mathrm{pred}},B_{\mathrm{gt}}) > 0.01
    \right).
\end{equation}
When the gate is closed, $R_{\mathrm{hier}}$ and $R_{\mathrm{bonus}}$ are disabled.
This prevents coordinate-level rewards from dominating invalid or non-overlapping predictions.

\paragraph{Tiered coordinate verification.}
$R_{\mathrm{hier}}$ measures coordinate accuracy at multiple tolerances.
Let $v_{\mathrm{pred}}^{(k)}$ and $v_{\mathrm{gt}}^{(k)}$ denote the reconstructed integer coordinates of the predicted and ground-truth boxes for dimension $k$.
We define:
\begin{equation}
\begin{aligned}
R_{\mathrm{hier}}
=
\frac{1}{4}
\sum_{k=1}^{4}
\sum_{l \in \{h,t,o\}}
\alpha_l
\mathbb{I}
\left(
\left|
v_{\mathrm{pred}}^{(k)}
-
v_{\mathrm{gt}}^{(k)}
\right|
< \tau_l
\right).
\end{aligned}
\end{equation}
Here, $l \in \{h,t,o\}$ denotes the hundreds, tens, and ones levels.
We set $\tau_h=50$, $\tau_t=10$, and $\tau_o=3$ on the 1000-bin scale, corresponding to error margins of $5\%$, $1\%$, and $0.3\%$.
The normalized weights are $(\alpha_h,\alpha_t,\alpha_o)=(1/6,1/3,1/2)$, giving $R_{\mathrm{hier}}\in[0,1]$.
All satisfied tolerance checks contribute to the sum.
These thresholds provide coarse-to-fine feedback for coordinate prediction.

\paragraph{Strict-IoU bonus.}
$R_{\mathrm{bonus}}$ rewards predictions that reach strict IoU milestones.
We use $\mathcal{M}=\{0.5,0.9,0.95\}$ as the milestone set:
\begin{equation}
    R_{\mathrm{bonus}}
    =
    \sum_{m \in \mathcal{M}}
    b_m \mathbb{I}(\mathrm{IoU} \geq m).
\end{equation}
The milestone weights are $(b_{0.5},b_{0.9},b_{0.95})=(0.5,1.0,2.0)$.
Bonuses accumulate, so a prediction with IoU $\geq0.95$ receives $R_{\mathrm{bonus}}=3.5$.
This term rewards both target localization and strict-threshold accuracy.

\section{Experiments}
\label{sec:experiments}

\subsection{Experimental Setup}
\label{subsec:setup}

\paragraph{Datasets and metrics.}
We evaluate our method on three widely used visual grounding benchmarks: RefCOCO~\cite{refcoco}, RefCOCO+~\cite{refcoco}, and RefCOCOg~\cite{refcocog}.
We report mean IoU (mIoU), Precision@0.5 (P@0.5), and Precision@0.95 (P@0.95).
P@0.5 measures coarse localization, while P@0.95 is used as the main strict-localization metric.
All values are reported as percentages.

\begin{table*}[t]
\centering
\caption{
Comparison with existing models on RefCOCO, RefCOCO+, and RefCOCOg.
All values are reported in percentage.
The best result in each column is shown in bold.
}
\label{tab:sota_comparison}
\setlength{\tabcolsep}{2.6pt}
\renewcommand{\arraystretch}{0.95}
\begin{tabular*}{\textwidth}{@{\extracolsep{\fill}}lccccccccc@{}}
\toprule
\multirow{2}{*}{Method}
& \multicolumn{3}{c}{RefCOCO}
& \multicolumn{3}{c}{RefCOCO+}
& \multicolumn{3}{c}{RefCOCOg} \\
\cmidrule(lr){2-4}
\cmidrule(lr){5-7}
\cmidrule(lr){8-10}
& mIoU & P@.5 & P@.95
& mIoU & P@.5 & P@.95
& mIoU & P@.5 & P@.95 \\
\midrule
\multicolumn{10}{@{}l}{\textit{Open-set detection models}} \\
OWLv2 & 41.5 & 40.2 & 10.1 & 37.3 & 35.1 & 7.1 & 30.2 & 29.2 & 6.4 \\
Grounding DINO & 56.2 & 57.5 & 25.7 & 56.7 & 57.2 & 23.7 & 58.8 & 59.8 & 24.7 \\
\midrule
\multicolumn{10}{@{}l}{\textit{Open-source VLMs}} \\
Qwen2.5-VL-3B & 55.6 & 60.2 & 3.8 & 52.6 & 62.3 & 2.1 & 49.8 & 44.1 & 3.2 \\
Qwen2.5-VL-7B & 60.7 & 67.7 & 6.1 & 58.2 & 64.8 & 5.6 & 50.6 & 53.4 & 5.9 \\
Qwen2.5-VL-72B & 62.5 & 70.4 & 8.1 & 58.9 & 66.1 & 7.3 & 55.1 & 59.7 & 7.9 \\
DeepSeek-VL2 & 51.1 & 56.2 & 1.1 & 44.6 & 46.6 & 0.7 & 38.8 & 34.2 & 0.3 \\
GLM-4.1V-9B & 83.5 & 91.6 & 28.9 & 80.2 & 87.6 & 27.8 & 80.1 & 83.6 & 26.3 \\
GLM-4.6V-106B & 82.0 & 88.5 & 32.9 & 75.6 & 80.9 & 32.8 & 80.2 & 86.2 & 34.4 \\
\midrule
\multicolumn{10}{@{}l}{\textit{Specialist VLMs based on Qwen2.5-VL-3B}} \\
VLM-R1 & 63.1 & 69.8 & 14.4 & 64.4 & 71.5 & 13.6 & 66.7 & 73.4 & 13.7 \\
Rex-Omni & 81.9 & 88.2 & 31.1 & 77.5 & 83.4 & \textbf{33.3} & 77.3 & 86.1 & 36.3 \\
Hi-R1 & \textbf{84.3} & \textbf{93.1} & \textbf{33.4}
& \textbf{81.8} & \textbf{89.9} & 32.9
& \textbf{80.3} & \textbf{86.4} & \textbf{39.4} \\
\bottomrule
\end{tabular*}
\end{table*}

\paragraph{Baselines.}
We compare with three groups of models.
The first group contains open-vocabulary detection models, including OWLv2~\cite{owlv2} and Grounding DINO~\cite{grounding-dino}.
The second group contains general-purpose VLMs, including Qwen2.5-VL~\cite{qwen2.5vl}, DeepSeek-VL2~\cite{deepseekvl2}, and GLM-family models~\cite{glm4.1,glm4.5}.
The third group contains specialist grounding VLMs, including VLM-R1~\cite{vlm-r1} and Rex-Omni~\cite{rex-omni}.
For evaluation, all predicted boxes are normalized and rescaled to the canonical $[0,999]$ coordinate space.

\paragraph{Implementation details.}
We use Qwen2.5-VL-3B~\cite{qwen2.5vl} as the base model.
Training follows a two-stage pipeline.
In the first stage, the model is fine-tuned with 80k RefCOCO training samples to adapt it to the Hi-Token output format.
In the second stage, the SFT model is further optimized with GRPO~\cite{grpo1,grpo2} on the same 80k training split using the Hi-GAR reward.
We denote the final model trained with both Hi-Token and Hi-GAR as \textbf{Hi-R1}.
Additional implementation details and hyperparameters are provided in the Appendix.

\begin{table}[t]
\centering
\caption{
Controlled representation and reward analysis on RefCOCO.
All values are percentages.
The upper block compares coordinate representations, including additional tuning for Flat; the lower block compares rewards with Hi-Token fixed.
}
\label{tab:controlled_analysis}
\setlength{\tabcolsep}{3.0pt}
\renewcommand{\arraystretch}{0.96}
\begin{tabular*}{\columnwidth}{@{\extracolsep{\fill}}lccc@{}}
\toprule
Setting & mIoU & P@.5 & P@.95 \\
\midrule
Flat SFT, fixed recipe & 58.3 & 64.1 & 23.0 \\
Flat SFT, extra tuning & 68.5 & 74.0 & 26.3 \\
Hi-Token SFT & 72.4 & 79.0 & 31.7 \\
\midrule
IoU-only GRPO & 76.1 & 89.0 & 30.3 \\
Full Hi-GAR, gate off & 80.4 & 91.9 & 31.3 \\
Full Hi-GAR, gate on & 84.3 & 93.1 & 33.4 \\
\bottomrule
\end{tabular*}
\end{table}

\subsection{Separating Representation and Refinement Gains}
\label{subsec:gain_attribution}

Table~\ref{tab:controlled_analysis} evaluates the complete coordinate encoding and the reward stage in two groups of comparisons.
With the backbone, 80k training samples, optimizer, schedule, decoding, and evaluator fixed, Hi-Token SFT improves RefCOCO P@0.95 from 23.0 to 31.7 over Flat SFT.
Additional Flat-specific tuning of the learning rate, schedule, and token-embedding initialization raises its P@0.95 to 26.3.
Hi-Token retains an advantage of 5.4 points over this best tested Flat configuration on RefCOCO, with corresponding advantages of 4.2 and 15.9 points on RefCOCO+ and RefCOCOg.
These gains are obtained before RL; full results appear in Appendix~\ref{appendix:controlled_results}.

\begin{figure*}[t]
    \centering
    \includegraphics[width=\textwidth]{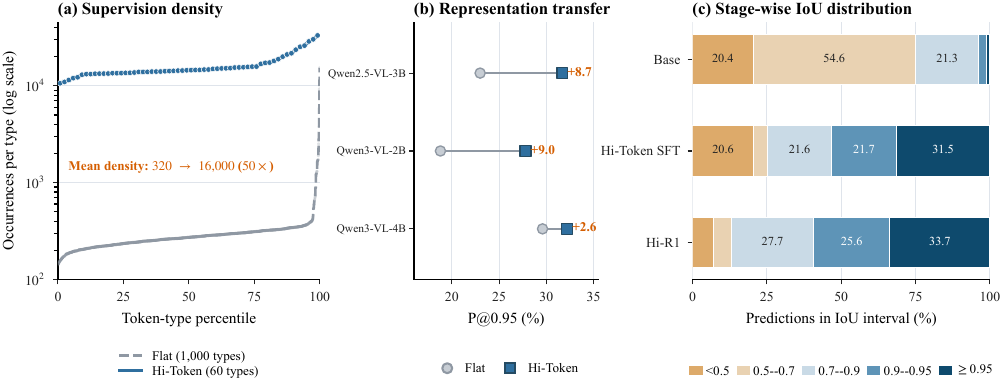}
    \caption{
    Hi-Token receives denser and more even supervision than Flat and improves strict localization across backbones; the stage-wise comparison shows the additional refinement from Hi-GAR.
    Panel (a) sorts token types by occurrence count within each representation and plots the complete distributions by percentile on a logarithmic vertical axis; the 50-fold mean-density difference follows from 16.7-fold fewer types and threefold more coordinate tokens per box.
    Panel (b) compares fixed-recipe Flat and Hi-Token SFT on RefCOCO across three backbones.
    Panel (c) reports the IoU distribution on RefCOCO for the Base model, Hi-Token SFT, and Hi-R1. The Base-to-SFT comparison includes supervised adaptation and the change in coordinate representation; the SFT-to-Hi-R1 comparison measures the GRPO stage with Hi-Token fixed.
    }
    \label{fig:gain_attribution}
\end{figure*}

Hi-Token combines explicit axis and place-value roles with frequent reuse of each coordinate-token type.
Across the same 80k paired examples, Flat emits four coordinate tokens per box over 1,000 types, whereas Hi-Token emits 12 tokens over 60 types.
These vocabulary and sequence sizes give the arithmetic mean supervision-density ratio $(12N/60)/(4N/1000)=50$ for $N$ training boxes.
Figure~\ref{fig:gain_attribution}(a) shows how the actual occurrences are distributed across token types.
The median count increases from 272 to 14,472; 985 of 1,000 Flat types occur fewer than 1,000 times, whereas every Hi-Token type occurs at least 10,615 times.
Counts are also more evenly distributed across Hi-Token types, as quantified in Appendix Table~\ref{tab:app_token_distribution}.
Counting distinct training examples gives 15,426 versus 318 examples per type on average, a 48.5-fold difference that measures coverage across examples.
The design preserves this broad coverage while assigning separate identities to digits with different axis and place-value roles.

Figure~\ref{fig:gain_attribution}(b) shows that the strict-metric benefit transfers beyond the original backbone.
Hi-Token improves RefCOCO P@0.95 by 8.7 points on Qwen2.5-VL-3B, 9.0 points on Qwen3-VL-2B, and 2.6 points on Qwen3-VL-4B under the same fixed-recipe comparison.
The effect is positive across all three models, although its magnitude varies.

Hi-GAR further improves localization with the coordinate representation fixed.
Relative to Hi-Token SFT, it improves mIoU by 11.9 points, P@0.5 by 14.1 points, and P@0.95 by 1.7 points.
IoU-only GRPO yields P@0.95 of 30.3, below the SFT value of 31.7, while full Hi-GAR reaches 33.4.
The validity gate contributes 3.9 mIoU points and 2.1 P@0.95 points over the gate-off configuration, which obtains 80.4 and 31.3, respectively.
Figure~\ref{fig:gain_attribution}(c) shows that Hi-GAR reduces
the proportion of predictions with IoU below 0.5 from 20.6\% to 7.1\%,
increases the proportion in $[0.7,0.95)$ from 43.3\% to 53.3\%,
and increases P@0.95 from 31.5\% to 33.7\%.
Together, the reward comparisons show that Hi-GAR improves overall box alignment while retaining and extending the strict-localization gains of Hi-Token SFT.

\paragraph{Role of SFT initialization.}
With 30k and 80k SFT samples, the models reach similar RefCOCO
P@0.5 after GRPO (93.0 and 93.1), but different P@0.95 scores
(6.5 and 33.4; Appendix Table~\ref{tab:app_dataset_scaling}).
Their P@0.95 scores before GRPO are 5.4 and 31.7, respectively.
These results highlight the importance of precise coordinate
learning during Hi-Token SFT for further refinement with Hi-GAR.

\subsection{Comparison with Existing Methods}
\label{subsec:main_results}

Table~\ref{tab:sota_comparison} compares Hi-R1 with existing detection models, general-purpose VLMs, and specialist grounding VLMs, which use different training regimes.
Table~\ref{tab:controlled_analysis} evaluates coordinate representations and rewards under matched training conditions.

Hi-R1 achieves strong performance on the RefCOCO family.
Compared with VLM-R1, Hi-R1 improves all reported metrics by a large margin.
Compared with Rex-Omni, Hi-R1 obtains higher mIoU and P@0.5 on all three datasets.
For strict localization, Hi-R1 improves P@0.95 on RefCOCO from 31.1 to 33.4 and on RefCOCOg from 36.3 to 39.4.
On RefCOCO+, Rex-Omni obtains a slightly higher P@0.95 score, 33.3 versus 32.9, while Hi-R1 still improves mIoU and P@0.5.
The results show that Hi-Token with Hi-GAR provides competitive or leading strict-localization performance while using a compact 3B backbone and 80k grounding samples.
The measured inference cost remains comparable to the SFT model because Hi-GAR is training-only; complete latency and time-per-token results are provided in Appendix~\ref{appendix:latency}.

\subsection{Boundary and Small-Object Diagnostics}
\label{subsec:failure_analysis}

\begin{figure*}[t]
    \centering
    \includegraphics[width=\textwidth]{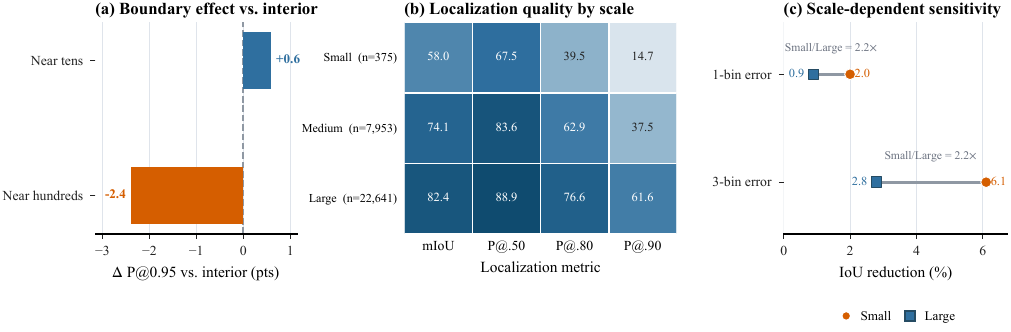}
    \caption{
    Hi-Token remains stable at tens transitions and retains useful small-object localization at moderate thresholds, while hundreds transitions and deterministic perturbations reveal localized geometric sensitivities.
    Panel (a) reports RefCOCO P@0.95 changes relative to the interior group (33.2) for two disjoint boundary groups.
    Panel (b) reports mIoU, P@0.5, P@0.8, and P@0.9 for Small ($<5\%$ GT area), Medium ($5\%$--$10\%$), and Large ($>10\%$) objects pooled over the five test splits; P@0.95 is reported in Appendix Table~\ref{tab:app_scale_aggregate}.
    Panel (c) pairs the IoU reductions for Small and Large boxes under deterministic one-bin and three-bin perturbations to every ground-truth coordinate; the ratio annotation divides the Small-object reduction by the Large-object reduction at the same perturbation.
    }
    \label{fig:boundary_scale_diagnostics}
\end{figure*}

We next examine localization around digit boundaries and across object scales.
For the digit-boundary analysis, coordinates near a hundreds boundary are assigned first, the remaining coordinates near a tens boundary are assigned second, and all other coordinates form the interior group.
Figure~\ref{fig:boundary_scale_diagnostics}(a) expresses each boundary group's P@0.95 as a change from the interior result of 33.2.
Near-tens coordinates differ by only $+0.6$ points, whereas near-hundreds coordinates drop by $2.4$ points and have a higher mean coordinate error of 44.1 bins, compared with 32.8 bins for interior coordinates.
The hundreds-boundary group includes image-edge cases, so its reported score covers both internal digit transitions and image edges.

The localization-quality matrix in Figure~\ref{fig:boundary_scale_diagnostics}(b) aggregates all five held-out splits, giving 375 Small, 7,953 Medium, and 22,641 Large examples.
For Small objects, Hi-R1 obtains mIoU of 58.0, P@0.5 of 67.5, P@0.8 of 39.5, and P@0.9 of 14.7.
These results show that the model often identifies the correct small-object region at moderate thresholds, although performance remains scale dependent as the localization criterion tightens.
Appendix Table~\ref{tab:app_scale_aggregate} reports P@0.95 for each object scale.
For Small objects, Hi-GAR increases P@0.95 from 1.07\% to 1.30\% (Table~\ref{tab:app_small_stage}).

Figure~\ref{fig:boundary_scale_diagnostics}(c) compares the IoU reductions caused by one-bin and three-bin coordinate perturbations for Small and Large boxes.
Perturbing each ground-truth coordinate by one bin reduces IoU by 2.0\% for Small objects and 0.9\% for Large objects; a three-bin perturbation reduces IoU by 6.1\% and 2.8\%, respectively.
Small boxes show larger IoU reductions at both perturbation levels.
Exact source values, per-dataset scale results, and additional qualitative cases are provided in Appendix~\ref{appendix:diagnostics}.

\section{Conclusion}
\label{sec:conclusion}
Generative visual grounding depends on reliable coordinate generation. We introduced Hi-Token, which represents each coordinate with axis-specific tokens for the hundreds, tens, and ones digits, providing coarse-to-fine structure and denser token supervision within the existing VLM architecture. Controlled experiments show improvements across the evaluated IoU range and three backbones. Hi-GAR further reduces low-overlap predictions through geometry-based training feedback and adds no inference-time component. Hi-R1 performs strongly against specialist models with latency close to that of the SFT model. These results show that coordinate representation is an important design choice for generative visual grounding.

\clearpage
\section*{Limitations}

Hi-Token operates in a fixed 1000-bin coordinate space. While this design provides a uniform representation across object scales, ultra-strict localization is more sensitive for small objects, motivating adaptive-resolution extensions. Our analyses identify denser token supervision and local coarse-to-fine structure as plausible contributors to the observed gains, but the current experiments do not isolate hierarchy, axis-specific vocabularies, and vocabulary size individually. Hi-GAR is evaluated with a fixed reward configuration and a validity-gate ablation; broader sensitivity studies are left to future work.

\section*{Ethical Considerations}

This work studies high-precision visual grounding on standard referring-expression benchmarks. Accurate grounding can support assistive perception, human--computer interaction, and fine-grained visual understanding. Incorrect grounding can also cause concrete harms: it may misidentify a person or object in surveillance, point a user to the wrong region in an assistive system, or propagate an incorrect localization into a safety-sensitive decision. The method may also be misused for tracking or privacy-sensitive monitoring. We do not intend it to be used for identifying, tracking, or profiling individuals without consent. Any deployment should use task-specific validation, follow applicable privacy regulations, obtain appropriate consent, and include safeguards against harmful or unauthorized use.

\bibliography{custom}

\clearpage
\appendix

\section{Training and Reproducibility Details}
\label{appendix:implementation}

\begin{table}[!htbp]
\centering
\caption{
Inference latency and time per token (TPT), in milliseconds.
}
\label{tab:app_latency}
\setlength{\tabcolsep}{2.6pt}
\renewcommand{\arraystretch}{0.95}
\begin{tabular*}{\columnwidth}{@{\extracolsep{\fill}}llcc@{}}
\toprule
Method & Dataset & Latency & TPT \\
\midrule
\multirow{3}{*}{Base}
& RefCOCO  & 9.08  & 0.37 \\
& RefCOCO+ & 10.50 & 0.37 \\
& RefCOCOg & 19.16 & 0.72 \\
\midrule
\multirow{3}{*}{Hi-Token SFT}
& RefCOCO  & 8.37  & 0.47 \\
& RefCOCO+ & 7.87  & 0.44 \\
& RefCOCOg & 18.61 & 1.03 \\
\midrule
\multirow{3}{*}{Hi-R1}
& RefCOCO  & 8.09  & 0.45 \\
& RefCOCO+ & 8.10  & 0.45 \\
& RefCOCOg & 19.58 & 1.09 \\
\bottomrule
\end{tabular*}
\end{table}

\begin{table*}[!htbp]
\centering
\caption{
Fixed-recipe representation comparison and additional Flat-specific tuning.
All values are percentages.
}
\label{tab:app_flat_tuning}
\setlength{\tabcolsep}{2.8pt}
\renewcommand{\arraystretch}{0.96}
\begin{tabular*}{\textwidth}{@{\extracolsep{\fill}}lccccc@{}}
\toprule
\multirow{2}{*}{Setting} & \multicolumn{3}{c}{RefCOCO} & \multicolumn{2}{c}{P@.95} \\
\cmidrule(lr){2-4}\cmidrule(lr){5-6}
& mIoU & P@.5 & P@.95 & RefCOCO+ & RefCOCOg \\
\midrule
Flat SFT, fixed recipe & 58.3 & 64.1 & 23.0 & 21.7 & 10.5 \\
Flat SFT, extra tuning & 68.5 & 74.0 & 26.3 & 25.6 & 21.1 \\
Hi-Token SFT, fixed recipe & 72.4 & 79.0 & 31.7 & 29.8 & 37.0 \\
\bottomrule
\end{tabular*}
\end{table*}

\begin{table}[!htbp]
\centering
\caption{
RefCOCO P@0.95 across backbones.
}
\label{tab:app_backbone_transfer}
\setlength{\tabcolsep}{2.5pt}
\renewcommand{\arraystretch}{0.96}
\begin{tabular*}{\columnwidth}{@{\extracolsep{\fill}}lccc@{}}
\toprule
Backbone & Flat & Hi-Token & Gain \\
\midrule
Qwen2.5-VL-3B & 23.0 & 31.7 & 8.7 \\
Qwen3-VL-2B & 18.8 & 27.8 & 9.0 \\
Qwen3-VL-4B & 29.6 & 32.2 & 2.6 \\
\bottomrule
\end{tabular*}
\end{table}

\subsection{Training Configuration}

All primary experiments use Qwen2.5-VL-3B-Instruct as the base model.
The model is first adapted to the target coordinate format through supervised fine-tuning on 80k RefCOCO training samples.
The GRPO stage is initialized from the SFT model and trained on the same 80k split with the Hi-GAR reward.

GRPO is implemented with the \texttt{verl}~\cite{verl} framework.
Training uses 4 nodes with 8 NVIDIA H800 GPUs per node, for a total of 32 GPUs.
We train for 2 epochs, corresponding to 156 optimization steps.
Each rollout batch contains 1,024 prompts, and the group size $G$ is 16 responses per prompt.
We use AdamW~\cite{adamw} with a learning rate of $5 \times 10^{-6}$, weight decay 0.01, and bfloat16 mixed precision.
The learning-rate schedule is constant with no warmup; the Adam coefficients are $(0.9,0.999)$, and the maximum gradient norm is 1.0.
Training responses are sampled at temperature 1.0 and top-$p$ 1.0.
The KL-loss coefficient is 0.01, using the low-variance KL estimator.
Policy updates use one PPO epoch, token-averaged loss, lower and upper clipping widths of 0.2 and 0.3, and a dual-clipping coefficient of 3.0.
Tensor parallelism is 4, and rollout-worker memory utilization is capped at 80\%.
The per-device microbatch sizes are 4 for policy updates and 8 for log-probability computation.
Prompt and response lengths are each capped at 2,048 tokens, and input images use a pixel budget between 262,144 and 409,600.
The vision tower is updated during training.
The full GRPO stage takes approximately 8 hours under this setup.

\subsection{Coordinate Formats and Baselines}

The fixed-recipe Flat baseline follows a Rex-Omni-style atomic location-token representation.
Each coordinate in $[0,999]$ is mapped to one of 1,000 dedicated location tokens.
The backbone, 80k training split, optimizer, schedule, decoding, and evaluator are the same as for fixed-recipe Hi-Token SFT.
This comparison evaluates the two coordinate encodings on 80k grounding samples.
Rex-Omni in Table~\ref{tab:sota_comparison} uses its published checkpoint trained on 22M multi-task samples; the grounding-only count is unspecified.

The Base row uses the pretrained Qwen2.5-VL-3B-Instruct model with its native numeric output.
The SFT rows measure adaptation to the corresponding coordinate format on the 80k grounding samples.

For a coordinate quantized to $478$, the Flat format uses one atomic token, \texttt{\textless 478\textgreater}.
Hi-Token instead uses the axis-specific sequence
\texttt{\textless x\_hundreds\_4\textgreater},
\texttt{\textless x\_tens\_7\textgreater}, and
\texttt{\textless x\_ones\_8\textgreater}.
The same value on the $y$ axis uses a separate token set.

\subsection{Reward Configuration and Prompt}

The reward weights are $\lambda_{\mathrm{fmt}}=0.5$, $\lambda_{\mathrm{iou}}=1.0$, $\lambda_{\mathrm{hier}}=0.5$, and $\lambda_{\mathrm{bonus}}=2.0$.
The coordinate tolerances are $\tau_h=50$, $\tau_t=10$, and $\tau_o=3$ on the 1,000-bin scale, corresponding to 5\%, 1\%, and 0.3\% errors. The corresponding normalized weights are $(\alpha_h,\alpha_t,\alpha_o)=(1/6,1/3,1/2)$.
The implementation adds 0.1, 0.2, and 0.3 for each satisfied tolerance and divides the sum over four coordinates by 2.4, which is equivalent to the normalized definition in Section~\ref{subsec:reward}.
The strict-IoU milestone set is $\mathcal{M}=\{0.5,0.9,0.95\}$, and the validity-gate threshold is 0.01 IoU.
Milestone weights are $(0.5,1.0,2.0)$ in ascending threshold order, and all reached milestones contribute cumulatively.
The tolerances were chosen as geometric margins independently of test-set performance.
Table~\ref{tab:controlled_analysis} compares the full reward with IoU-only training and removal of the validity gate at these settings.

SFT and GRPO use the same prompt:
\begin{quote}
\texttt{Please provide the bounding box coordinate of the region this sentence describes: <expr>.}
\end{quote}
The placeholder \texttt{<expr>} is replaced by the referring expression.
During reward computation, we set \texttt{skip\_special\_tokens=False} so that the axis-specific coordinate tokens remain available to the parser.

\subsection{Evaluation Protocol}

Predicted boxes are converted to the normalized $[0,999]$ coordinate space.
For Hi-Token, the parser reconstructs each integer coordinate from the axis-specific hundreds, tens, and ones tokens.
For Flat outputs, each location token maps directly to its integer coordinate.
The reconstructed boxes are evaluated with IoU.
Model-specific parsers convert each output format to bounding-box coordinates, which are scored with the same IoU evaluator.
Object-scale groups use ground-truth area on the normalized $1000\times1000$ canvas: Small is $<5\%$, Medium is $5\%$--$10\%$, and Large is $>10\%$.

\subsection{Inference Latency}
\label{appendix:latency}

Latency is measured with the same decoding configuration and vLLM~\cite{vllm} backend.
The base tokenizer already splits a three-digit numeric coordinate into three digit tokens.
Hi-Token also uses three tokens for a three-digit coordinate; extra output tokens mainly arise for one- or two-digit coordinates that are padded to hundreds, tens, and ones.
Hi-GAR is training-only and adds no inference-time module.

\begin{table*}[t]
\centering
\caption{
Coordinate-token supervision budget and distinct-example coverage on the 80k training split.
}
\label{tab:app_token_density}
\setlength{\tabcolsep}{4.0pt}
\renewcommand{\arraystretch}{0.96}
\begin{tabular*}{\textwidth}{@{\extracolsep{\fill}}lrrrrr@{}}
\toprule
Representation & Types & Tok./box & Total & Avg./type & Avg. ex./type \\
\midrule
Flat atomic & 1,000 & 4 & 320k & 320 & 318 \\
Hi-Token & 60 & 12 & 960k & 16,000 & 15,426 \\
\bottomrule
\end{tabular*}
\end{table*}

\section{Additional Controlled Results}
\label{appendix:controlled_results}

\subsection{Additional Flat-Baseline Tuning}

Table~\ref{tab:app_flat_tuning} reports fixed-recipe SFT and additional Flat-specific tuning of the learning rate, schedule, and token-embedding initialization.
The best tested Flat configuration improves RefCOCO mIoU from 58.3 to 68.5, P@0.5 from 64.1 to 74.0, and P@0.95 from 23.0 to 26.3.
Hi-Token with the fixed recipe reaches 72.4, 79.0, and 31.7, respectively.
Its P@0.95 advantages over the additionally tuned Flat model are 5.4 points on RefCOCO, 4.2 on RefCOCO+, and 15.9 on RefCOCOg.

\subsection{Backbone Transfer and Supervision Density}

Table~\ref{tab:app_backbone_transfer} provides the exact values plotted in Figure~\ref{fig:gain_attribution}(b).
All rows use the same fixed-recipe Flat-versus-Hi SFT comparison for the corresponding backbone.
The effect is positive for all three models, although the gain is smaller for Qwen3-VL-4B.

Tables~\ref{tab:app_token_density} and~\ref{tab:app_token_distribution} report the token-frequency diagnostic on the same 80k paired training examples.
The paired representations contain the same sample identifiers, images, queries, and decoded coordinates; all 160,000 encoded examples parse successfully.
Every coordinate token is observed in both representations.

\begin{table*}[t]
\centering
\caption{
Distribution of occurrence counts across coordinate-token types.
All statistics use the complete vocabulary; $H/\log|\mathcal{V}|$ is normalized entropy.
}
\label{tab:app_token_distribution}
\setlength{\tabcolsep}{4.0pt}
\renewcommand{\arraystretch}{0.96}
\begin{tabular*}{\textwidth}{@{\extracolsep{\fill}}lrrrrrrrr@{}}
\toprule
Representation & Min. & Q1 & Median & Q3 & Max. & CV & Gini & $H/\log|\mathcal{V}|$ \\
\midrule
Flat atomic & 142 & 235 & 272 & 312 & 15,293 & 1.844 & 0.237 & 0.954 \\
Hi-Token & 10,615 & 13,560 & 14,472 & 15,753 & 33,052 & 0.284 & 0.130 & 0.991 \\
\bottomrule
\end{tabular*}
\end{table*}

The exact 50-fold mean ratio follows from a coordinate vocabulary that is 16.7 times smaller than Flat and three times as many supervised coordinate tokens per box.
The distinct-example ratio is 48.5-fold, quantifying coverage across training examples.
The Gini coefficient measures inequality in occurrence counts, with lower values indicating more even coverage; normalized entropy increases as the distribution becomes more uniform.
Hi-Token has a lower Gini coefficient (0.130 versus 0.237) and higher normalized entropy (0.991 versus 0.954).
These statistics characterize the reuse and coverage of the complete coordinate encoding.

\begin{table*}[!htbp]
\centering
\caption{
Source summary for Figure~\ref{fig:reward_curve}.
Delta is the last-10-step mean minus the first-10-step mean.
}
\label{tab:app_reward_summary}
\setlength{\tabcolsep}{4.5pt}
\renewcommand{\arraystretch}{0.96}
\begin{tabular*}{\textwidth}{@{\extracolsep{\fill}}lrrrrr@{}}
\toprule
Series & Step 1 & Step 156 & First 10 mean & Last 10 mean & Delta \\
\midrule
Overall & 6.2218 & 6.8186 & 6.3696 & 6.7044 & 0.3348 \\
Format & 0.4834 & 0.4987 & 0.4948 & 0.4986 & 0.0038 \\
IoU & 0.7557 & 0.8117 & 0.7799 & 0.8074 & 0.0275 \\
Hierarchical & 0.6531 & 0.7541 & 0.6689 & 0.7461 & 0.0773 \\
Strict-IoU bonus & 2.3281 & 2.5655 & 2.3802 & 2.5127 & 0.1324 \\
\bottomrule
\end{tabular*}
\end{table*}

\subsection{SFT Initialization and Data Scale}

Table~\ref{tab:app_dataset_scaling} reports localization performance
before and after GRPO for models initialized with 30k and 80k SFT samples across the RefCOCO family.

\begin{table*}[!htbp]
\centering
\caption{
SFT initialization and GRPO results across the RefCOCO family.
All values are percentages.
}
\label{tab:app_dataset_scaling}
\setlength{\tabcolsep}{2.6pt}
\renewcommand{\arraystretch}{0.95}
\begin{tabular*}{\textwidth}{@{\extracolsep{\fill}}lccccccccc@{}}
\toprule
\multirow{2}{*}{Setting}
& \multicolumn{3}{c}{RefCOCO}
& \multicolumn{3}{c}{RefCOCO+}
& \multicolumn{3}{c}{RefCOCOg} \\
\cmidrule(lr){2-4}\cmidrule(lr){5-7}\cmidrule(lr){8-10}
& mIoU & P@.5 & P@.95 & mIoU & P@.5 & P@.95 & mIoU & P@.5 & P@.95 \\
\midrule
SFT, 30k initialization & 66.0 & 77.9 & 5.4 & 63.6 & 74.8 & 5.3 & 62.1 & 73.6 & 4.3 \\
GRPO from 30k SFT & 78.6 & 93.0 & 6.5 & 75.7 & 89.1 & 5.8 & 73.3 & 86.2 & 5.9 \\
SFT, 80k initialization & 72.4 & 79.0 & 31.7 & 70.2 & 76.6 & 29.8 & 70.8 & 75.8 & 37.0 \\
GRPO from 80k SFT & 84.3 & 93.1 & 33.4 & 81.8 & 89.9 & 32.9 & 80.3 & 86.4 & 39.4 \\
\bottomrule
\end{tabular*}
\end{table*}

\section{Diagnostic Source Tables}
\label{appendix:diagnostics}

\subsection{Stage-Wise IoU Distribution}

Table~\ref{tab:app_iou_distribution} reports the values shown in
Figure~\ref{fig:gain_attribution}(c) for the same RefCOCO examples.
The Base-to-SFT comparison includes supervised adaptation and
the change in coordinate representation.
The SFT-to-Hi-R1 comparison measures the GRPO stage with Hi-Token fixed.

\begin{table*}[!htbp]
\centering
\caption{
Percentage of predictions in each IoU interval on RefCOCO 
($N=5{,}657$ for each model).
Rows sum to 100\% up to rounding.
}
\label{tab:app_iou_distribution}
\setlength{\tabcolsep}{4.0pt}
\renewcommand{\arraystretch}{0.96}
\begin{tabular*}{\textwidth}{@{\extracolsep{\fill}}lccccc@{}}
\toprule
Stage & IoU $<0.5$ & $[0.5,0.7)$ & $[0.7,0.9)$
& $[0.9,0.95)$ & IoU $\geq0.95$ \\
\midrule
Base & 20.4 & 54.6 & 21.3 & 2.5 & 1.2 \\
Hi-Token SFT & 20.6 & 4.6 & 21.6 & 21.7 & 31.5 \\
Hi-R1 & 7.1 & 6.0 & 27.7 & 25.6 & 33.7 \\
\bottomrule
\end{tabular*}
\end{table*}

\subsection{Digit-Boundary Analysis}

Table~\ref{tab:app_rollover} reports the complete source values for Figure~\ref{fig:boundary_scale_diagnostics}(a).
Groups are disjoint: hundreds-boundary cases are assigned first, then the remaining tens-boundary cases, followed by interior coordinates.
The hundreds group also includes edge coordinates, so the observed reduction cannot be attributed only to digit rollover.

\begin{table}[!htbp]
\centering
\caption{
RefCOCO performance by coordinate-boundary group.
Mean error is measured in bins on the 1,000-bin scale.
}
\label{tab:app_rollover}
\setlength{\tabcolsep}{2.0pt}
\renewcommand{\arraystretch}{0.96}
\begin{tabular*}{\columnwidth}{@{\extracolsep{\fill}}lrrrr@{}}
\toprule
Group & Ratio & Mean err. & mIoU & P@.95 \\
\midrule
Near tens & 32.6 & 31.1 & 80.9 & 33.8 \\
Near hundreds & 14.7 & 44.1 & 77.2 & 30.8 \\
Interior & 52.7 & 32.8 & 80.2 & 33.2 \\
\bottomrule
\end{tabular*}
\end{table}

\subsection{Object Scale and Coordinate Sensitivity}

Table~\ref{tab:app_scale_aggregate} provides the complete source values for Figure~\ref{fig:boundary_scale_diagnostics}(b).
The five held-out splits contain 30,969 examples in total.
Intervals are defined using ground-truth area on the normalized canvas.

\begin{table*}[!htbp]
\centering
\caption{
Five-split performance by object scale.
All metrics except $N$ are percentages.
}
\label{tab:app_scale_aggregate}
\setlength{\tabcolsep}{4.0pt}
\renewcommand{\arraystretch}{0.96}
\begin{tabular*}{\textwidth}{@{\extracolsep{\fill}}lrrrrrrr@{}}
\toprule
Scale & $N$ & mIoU & P@.5 & P@.7 & P@.8 & P@.9 & P@.95 \\
\midrule
Small & 375 & 58.0 & 67.5 & 51.2 & 39.5 & 14.7 & 1.30 \\
Medium & 7,953 & 74.1 & 83.6 & 73.1 & 62.9 & 37.5 & 17.5 \\
Large & 22,641 & 82.4 & 88.9 & 83.0 & 76.6 & 61.6 & 39.0 \\
\bottomrule
\end{tabular*}
\end{table*}

The training-stage comparison in Table~\ref{tab:app_small_stage} shows that Hi-GAR improves Small-object localization across all reported metrics, with P@0.95 increasing from 1.07\% to 1.30\%.

\begin{table}[!htbp]
\centering
\caption{
Small-object performance across training stages. All values are percentages.
}
\label{tab:app_small_stage}
\setlength{\tabcolsep}{2.5pt}
\renewcommand{\arraystretch}{0.96}
\begin{tabular*}{\columnwidth}{@{\extracolsep{\fill}}lrrrr@{}}
\toprule
Setting & mIoU & P@.5 & P@.9 & P@.95 \\
\midrule
Hi-Token SFT & 51.5 & 61.3 & 13.6 & 1.07 \\
Hi-R1 & 58.0 & 67.5 & 14.7 & 1.30 \\
\bottomrule
\end{tabular*}
\end{table}

Table~\ref{tab:app_perturbation} reports the deterministic source values for Figure~\ref{fig:boundary_scale_diagnostics}(c).
Each ground-truth coordinate is perturbed by the stated number of bins.
These values quantify box-overlap sensitivity to ground-truth coordinate perturbations.

\begin{table}[!htbp]
\centering
\caption{
IoU reduction under coordinate perturbations, in percentage.
}
\label{tab:app_perturbation}
\setlength{\tabcolsep}{3.0pt}
\renewcommand{\arraystretch}{0.96}
\begin{tabular*}{\columnwidth}{@{\extracolsep{\fill}}lrr@{}}
\toprule
Scale & 1 bin & 3 bins \\
\midrule
Small & 2.0 & 6.1 \\
Large & 0.9 & 2.8 \\
\bottomrule
\end{tabular*}
\end{table}

Table~\ref{tab:app_scale_per_dataset} reports the scale analysis for each dataset.
Table~\ref{tab:app_scale_aggregate} provides the five-split aggregate used in the main discussion.

\begin{table}[!htbp]
\centering
\caption{
Per-dataset performance by object scale.
Ratio is the percentage of samples in each dataset group.
}
\label{tab:app_scale_per_dataset}
\setlength{\tabcolsep}{2.3pt}
\renewcommand{\arraystretch}{0.93}
\begin{tabular*}{\columnwidth}{@{\extracolsep{\fill}}llrrrr@{}}
\toprule
Dataset & Scale & Ratio & mIoU & P@.5 & P@.95 \\
\midrule
\multirow{4}{*}{RefCOCO}
& Small & 1.0 & 66.1 & 72.7 & 0.0 \\
& Medium & 19.9 & 79.0 & 90.0 & 16.5 \\
& Large & 79.1 & 86.1 & 94.1 & 37.6 \\
& All & 100.0 & 84.5 & 93.1 & 33.0 \\
\midrule
\multirow{4}{*}{RefCOCO+}
& Small & 1.0 & 62.7 & 72.4 & 0.0 \\
& Medium & 19.6 & 76.0 & 86.9 & 16.7 \\
& Large & 79.3 & 83.4 & 90.9 & 37.3 \\
& All & 100.0 & 81.7 & 89.9 & 32.9 \\
\midrule
\multirow{4}{*}{RefCOCOg}
& Small & 0.1 & 71.2 & 85.7 & 7.1 \\
& Medium & 32.1 & 73.6 & 82.6 & 19.8 \\
& Large & 67.7 & 83.2 & 88.1 & 48.9 \\
& All & 100.0 & 80.1 & 86.3 & 39.5 \\
\bottomrule
\end{tabular*}
\end{table}

\section{Hi-GAR Training Dynamics}
\label{appendix:reward_dynamics}

\begin{figure*}[!htbp]
    \centering
    \includegraphics[width=\textwidth]{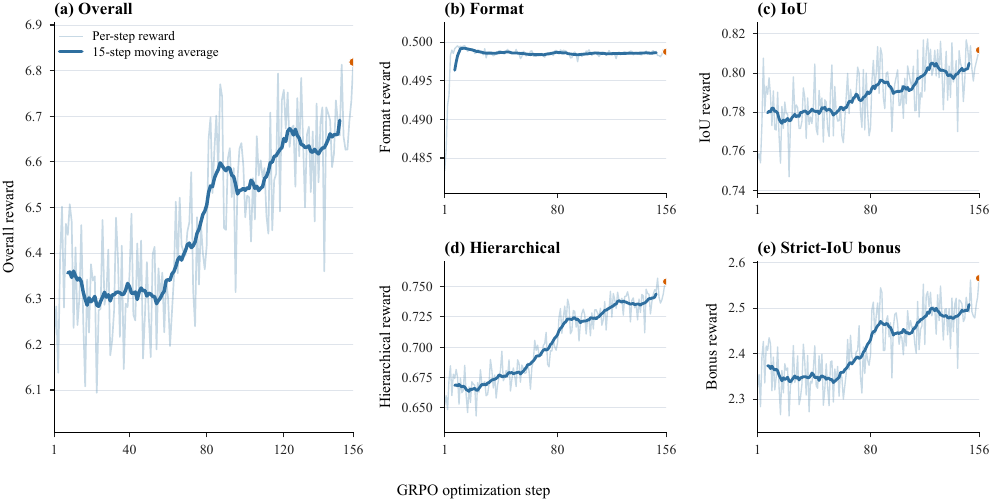}
    \caption{
    The geometric components of Hi-GAR improve over training, while format compliance stabilizes during the first few steps.
    Light lines show all 156 logged per-step batch-average rewards from one GRPO run, dark lines show a descriptive 15-step moving average, and orange markers denote the final logged values.
    The panels use separate vertical scales because the reward components have different numerical ranges.
    }
    \label{fig:reward_curve}
\end{figure*}

Figure~\ref{fig:reward_curve} shows the reward components over the 156 GRPO optimization steps.
The overall reward rises over the run, and the IoU, hierarchical, and strict-IoU bonus components show corresponding upward trends.
The format component approaches its plateau early, indicating that later changes are driven mainly by geometric quality rather than output validity.

Table~\ref{tab:app_reward_summary} summarizes the exact raw series.
Each series contains one logged batch-average reward per optimization step.
The first-window and last-window columns are arithmetic means over steps 1--10 and 147--156, respectively.
The logged Format value includes its outer weight of 0.5; Hierarchical and Strict-IoU bonus are logged before applying their outer weights of 0.5 and 2.0.
Overall is the sum of the weighted components.

\section{Additional Qualitative Results}
\label{appendix:qualitative}

\begin{figure}[!htbp]
  \centering
  \includegraphics[width=\linewidth, trim=10 30 375 10, clip]{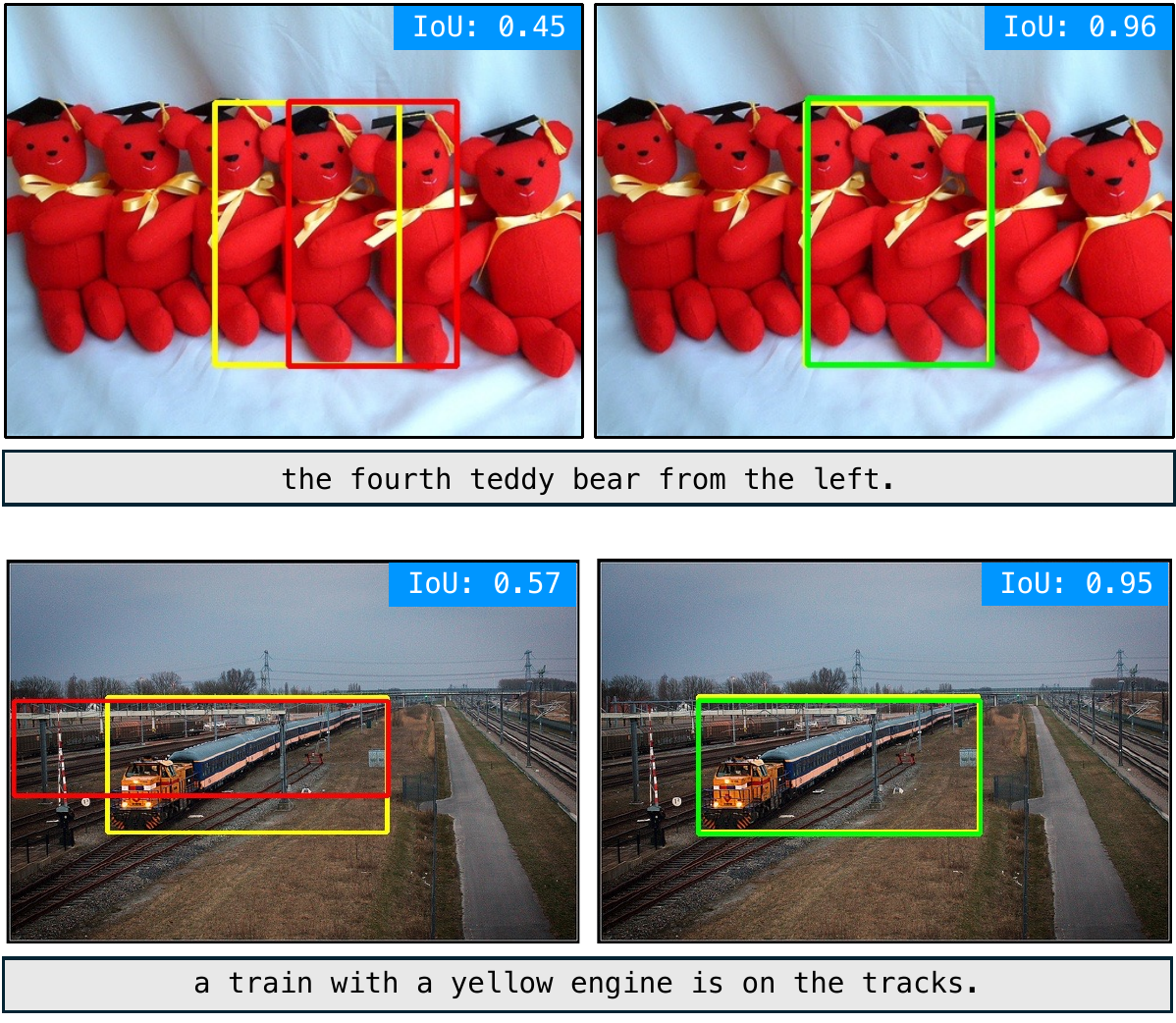}
  \caption{
  Qualitative effect of geometry-aware post-training.
  The SFT-only Hi-Token model covers the target but has boundary errors, whereas Hi-R1 produces tighter boxes in these examples.
  }
  \label{fig:qualitative_rl_impact}
\end{figure}

Figure~\ref{fig:qualitative_rl_impact} illustrates boundary refinement after Hi-GAR.
The examples illustrate boundary changes in individual predictions; Figure~\ref{fig:gain_attribution}(c) reports the corresponding distribution across evaluation examples.

\begin{figure}[!htbp]
    \centering
    \includegraphics[
        width=\linewidth,
        trim=10 130 370 10,
        clip
    ]{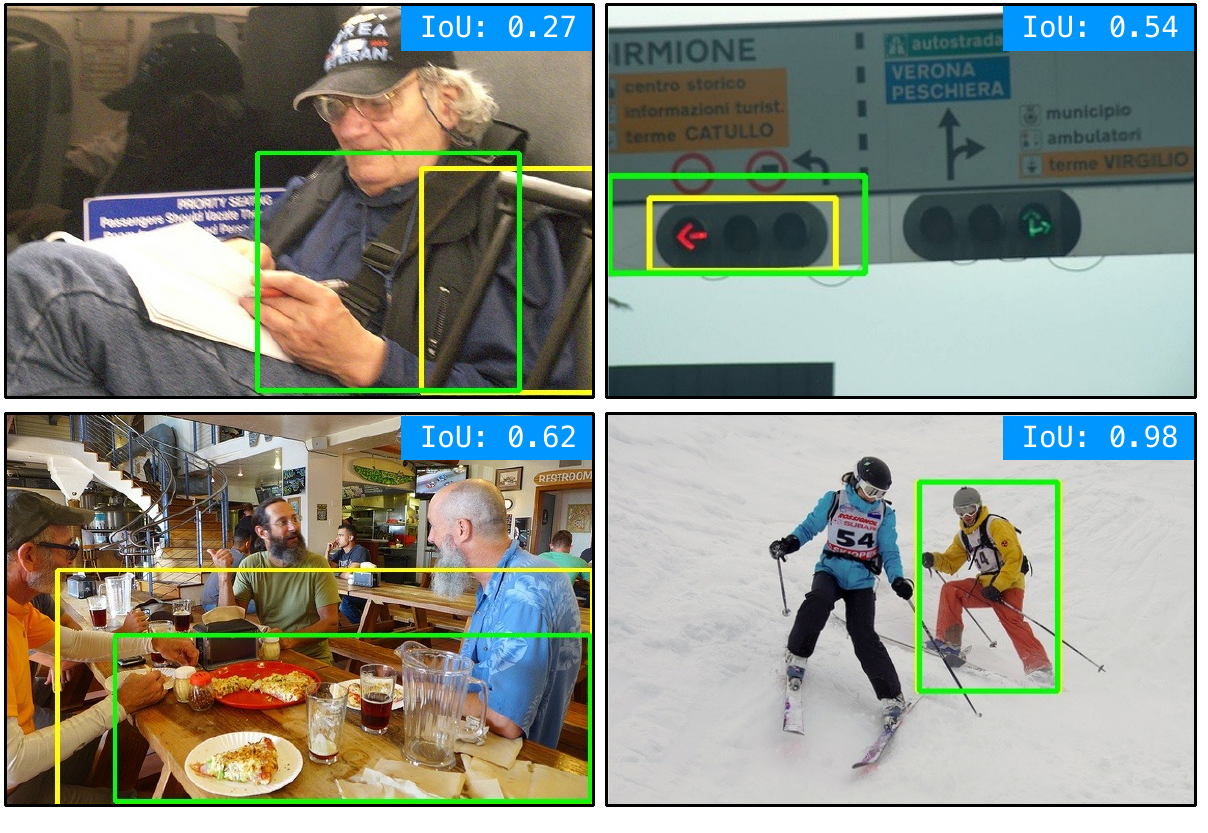}
    \caption{
    Additional Hi-R1 predictions at different IoU levels.
    The examples include both accurate predictions and representative failure cases.
    }
    \label{fig:comprehensive_cases}
\end{figure}

Figure~\ref{fig:comprehensive_cases} contains additional Hi-R1 predictions over a range of IoU values.
Low-IoU examples include ambiguous boundaries, part-versus-whole references, and genuine localization errors.

\section{Artifact Use and Licenses}
\label{appendix:artifacts}

This work uses publicly available research artifacts, including Qwen2.5-VL-3B~\cite{qwen2.5vl}, the RefCOCO benchmarks~\cite{refcoco,refcocog}, \texttt{verl}~\cite{verl}, and vLLM~\cite{vllm}.
We use official referring-expression grounding benchmarks based on natural images and English expressions, without introducing new datasets, human annotations, or redistributing the original data.
Released code, checkpoints, and derived artifacts should follow the licenses of the underlying datasets, models, and software.

\end{document}